\documentclass{article}
\usepackage{ijcai19}

\usepackage{times}
\usepackage{soul}
\usepackage{url}
\usepackage[hidelinks]{hyperref}
\usepackage[utf8]{inputenc}
\usepackage[small]{caption}
\usepackage{graphicx}
\usepackage{multirow}
\usepackage{amsmath}
\usepackage{booktabs}
\usepackage{algorithm}
\usepackage{algorithmic}
\usepackage{color}
\usepackage{amssymb}
\usepackage{amsbsy}

\usepackage[dvipsnames]{xcolor}

\title{Dynamic Feature Fusion for Semantic Edge Detection}

\author{
Yuan Hu$^{1,2}$
\and
Yunpeng Chen$^3$\and
Xiang Li$^{1,2}$\And
Jiashi Feng$^3$
\affiliations
$^1$Institute of Remote Sensing and Digital Earth, CAS, Beijing 100094, China\\
$^2$University of Chinese Academy of Sciences, Beijing 100049, China\\
$^3$National University of Singapore
\emails
\{huyuan, lixiang01\}@radi.ac.cn,
chenyunpeng@u.nus.edu,
elefjia@nus.edu.sg
}
\begin{document}

\maketitle

\begin{abstract}
  Features from multiple scales can greatly benefit the semantic edge detection task if they are well fused. However, the prevalent semantic edge detection methods apply a fixed weight fusion strategy where images with different semantics are forced to share the same weights, resulting in universal fusion weights for all images and locations regardless of their different semantics or local context.
  In this work, we propose a novel dynamic feature fusion strategy that assigns different fusion weights for different input images and locations adaptively. This is achieved by a proposed weight learner to infer proper fusion weights over multi-level features for each location of the feature map, conditioned on the specific input. In this way, the heterogeneity in contributions made by different locations of feature maps and input images can be better considered and thus help produce more accurate and sharper edge predictions.
  We show that our model with the novel dynamic feature fusion is superior to fixed weight fusion and also the na\"ive location-invariant weight fusion methods, via comprehensive experiments on benchmarks Cityscapes and SBD. In particular, our method outperforms all existing well established methods and achieves new state-of-the-art.
\end{abstract}

\section{Introduction}

\begin{figure}[t]
\centering
\includegraphics[width=.5\textwidth]{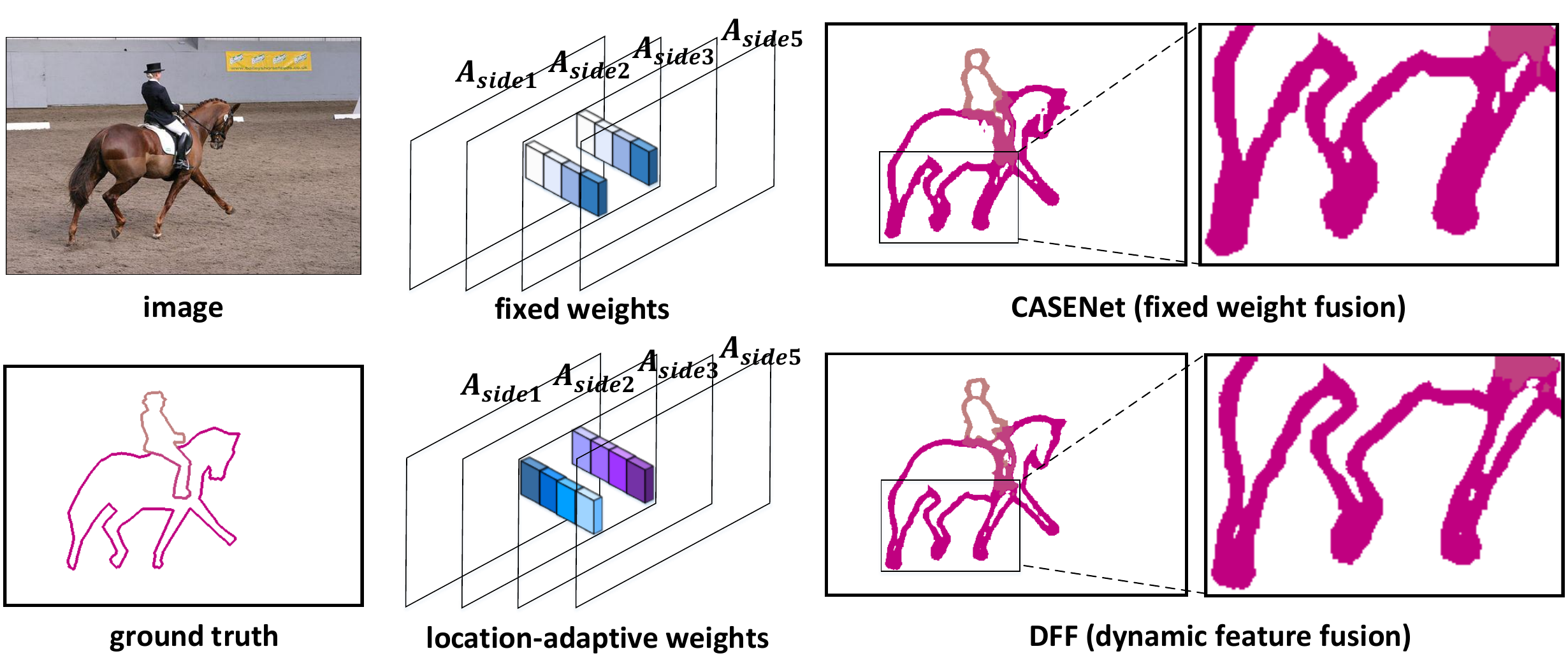}
\caption{Illustration of fixed weight fusion and proposed dynamic feature fusion (DFF). The proposed model actively learns customized fusion weights for each location, to fuse low-level features ($A_{side1} \sim A_{side3}$) and a high-level feature ($A_{side5}$). In this way, sharper  boundaries are produced through dynamic feature fusion compared with CASENet (fixed weight fusion model).}
\vspace{-4mm}
\label{fig1}
\end{figure}

The task of semantic edge detection (SED) is aimed at both detecting visually salient edges and recognizing their categories, or more concretely, locating fine  edges utlizing low-level features and meanwhile identifying semantic categories with abstracted high-level features. An intuitive way for a deep CNN model to achieve both targets is to integrate high-level semantic features with low-level category-agnostic edge features via a fusion model, which  is conventionally designed following a fixed weight fusion strategy, independent of the input, as illustrated in the top row in Figure~\ref{fig1}.

In many existing deep SED models~\cite{yu2017casenet,liu2018semantic,yu2018simultaneous}, fixed weight fusion of multi-level features  is implemented through $1 \times 1$ convolution, where the learned convolution kernel serves as the fusion weights. However, this fusion strategy cannot fully exploit multi-level information, especially the low-level features. This is because, first, it applies the same fusion weights to all the input images and ignores their variations in contents, illumination, \emph{etc}. The distinct properties of a specific input need be treated adaptively for revealing the subtle edge details. Besides, for a same input image, different spatial locations on the corresponding feature map convey different information, but the fixed weight fusion manner applies the same weights to all these locations, regardless of their different semantic classes or object parts. This would unfavorably  force the model to learn universal fusion weights for all the categories and locations. Consequently, a bias would be caused toward high-level features, and the power of multi-level response fusion is significantly weakened.

In this work, we propose a \textit{Dynamic Feature Fusion }(DFF) method that assigns adaptive fusion weights to each location individually, aiming to generate a fused edge map adaptive to the specific content of each image, as illustrated in the bottom row in Figure~\ref{fig1}. In particular, we design a novel location-adaptive weight learner that actively learns customized location-specific fusion weights conditional on the feature map content for multi-level response maps. As shown in Figure~\ref{fig1}, low-level features ($A_{side1} \sim A_{side3}$) and a high-level feature ($A_{side5}$) are merged to produce the final fused output. Low-level feature maps give high response on fine details, such as the edges inside objects, whereas high-level ones are coarser and only exhibit strong response at the object-level boundaries. This location-adaptive weight learner tailors fusion weights for each {individual location}. For example, for the boundaries of the horse, fusion weights are biased toward low-level features to fully take advantage of the accurate located edges. For the interior of the horse, higher weights are assigned to high-level features to suppress the fragmentary and trivial edge responses inside the object.

The proposed DFF model consists of two main components: a \textit{feature extractor with a normalizer} and an \textit{adaptive weight fusion module}. The feature extractor primarily scales the multi-level responses to the same magnitude,  preparing for the down-streaming fusion operation. The adaptive weight fusion module performs following two computations. First, it dynamically generates location-specific fusion weights conditioned  on the image content. Then, the location-specific fusion weights are applied to actively fuse the high-level and low-level response maps. The adaptive weight fusion module is capable of fully excavating the potentialities of multi-level responses, especially the low-level ones, to produce better fusion output for every single location.

In summary, our main contributions are:
\begin{itemize}
\setlength\itemsep{0em}
\item {For the first time, this work reveals  limitations of the popular fixed weight fusion for SED, and explains why it does not produce satisfactory fusion results as expected.
}
\item We propose a dynamic feature fusion (DFF) model. To our best knowledge, it is the first work to learn adaptive fusion weights conditioned on input contents to merge multi-level features in the research field of SED.
\item The proposed DFF model achieves new state-of-the-art on the SED task.
\end{itemize}

\section{Related Work}

%\paragraph{Semantic edge detection.}
The task of category-aware semantic edge detection was first introduced by~\cite{hariharan2011semantic}. It is tackled usually as a multi-class problem~\cite{hariharan2011semantic,bertasius2015high,bertasius2016semantic,maninis2016convolutional} at first, in which only one semantic class is associated with each located boundary pixel.  From CASENet~\cite{yu2017casenet}, researchers begin to address this task as a multi-label problem where each edge pixel can be associated with more than one semantic class simultaneously. Recently, deep learning based models, such as SEAL~\cite{yu2018simultaneous} and DDS~\cite{liu2018semantic}, further lift the performance of semantic edge detection to new state-of-the-art.

%\paragraph{Fixed weight fusion in edge detection.}
The tradition of employing fixed weight fusion can be pinpointed to HED~\cite{xie2015holistically}, in which a weighted-fusion layer (implemented by $1 \times 1$ convolution layer) is employed to merge the side-outputs. RCF~\cite{liu2017richer} and RDS~\cite{liu2016learning} follow this simple strategy to perform category-agnostic edge detection. CASENet~\cite{yu2017casenet}, SEAL~\cite{yu2018simultaneous} and DDS~\cite{liu2018semantic} extend the approach with a K-grouped 1 $\times$ 1 convolution layer to generate the K-channel fused activation map. In this paper, we demonstrate the above fixed fusion strategy cannot sufficiently leverage multi-scale responses for producing better fused output. Instead, our proposed adaptive weight fusion module enables the network to actively learn the location-aware fusion weights conditioned on the individual input content.

\begin{figure*}[t]
\centering
\resizebox{\textwidth}{!}{
\includegraphics[]{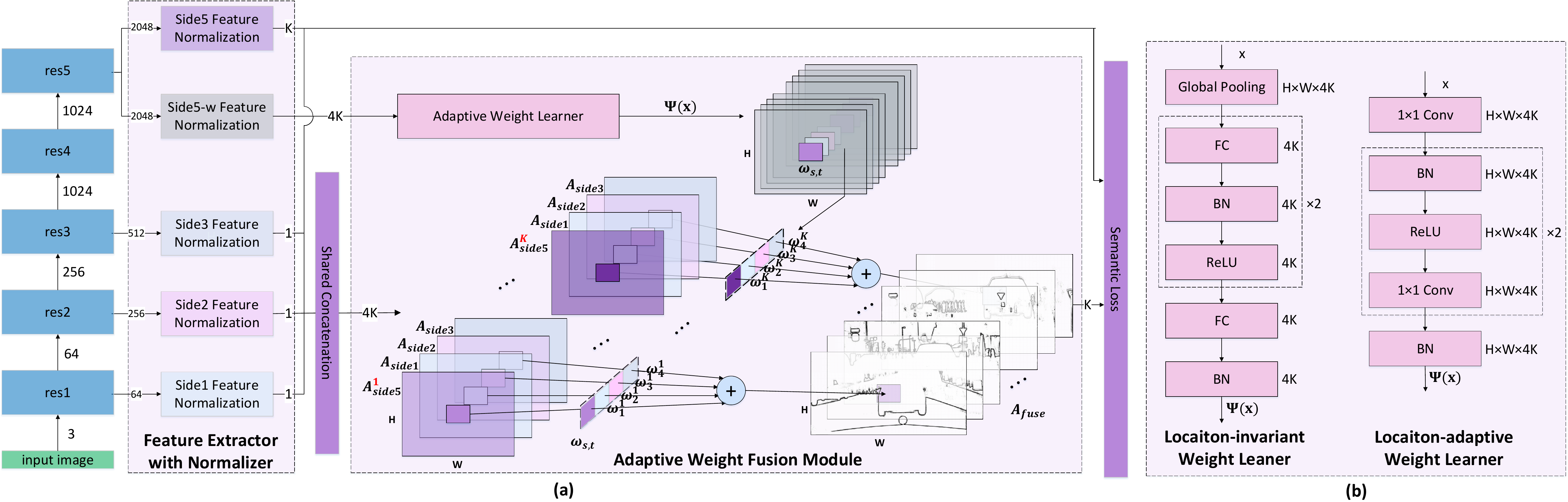}
}
\vspace{-5mm}
\caption{Overall architecture of our model. (a) The input image is fed into the ResNet backbone to generate a set of features (response maps) with different scales. Side feature normalization blocks are connected to the first three and the fifth stack of residual blocks to produce Side1-3 and Side5 response maps with the same response magnitude. Shared concatenation (Eqn. (1) and (2)) is then applied to concatenate Side1-3 and Side5. The \emph{Side5-w feature normalization} block followed by a \emph{location-adaptive weight learner} form another branch extended from res5 to predict dynamic location-aware fusion weights $\Psi(x)$. Then element-wise multiplication and category-wise summation are applied to the location-aware fusion weights $\Psi(x)$ and the concatenated response map $A_{cat}$ to generate the final fused output $A_{fuse}$. The semantic loss is employed to supervise $A_{side5}$ and the final fused output $A_{fuse}$. (b) \emph{Location-invariant weight learner} and \emph{location-adaptive weight learner} take as input the feature map $x$ and output location-invariant fusion weights and location-adaptive fusion weights $\Psi(x)$ respectively.}
\vspace{-4mm}
\label{fig2}
\end{figure*}

 Multi-level representations within a convolutional neural network have been shown effective in many vision tasks, such as object detection~\cite{liu2016ssd,lin2017feature}, semantic segmentation~\cite{long2015fully,yu2015multi}, boundary detection~\cite{xie2015holistically,liu2017richer}. Responses from different stages of a network tend to vary significantly in magnitude. Scaling the multi-scale activations to a similar magnitude would benefit the following prediction or fusion consistently. For example, SSD~\cite{liu2016ssd} performs L2 normalization on low-level feature maps before making multi-scale predictions.~\cite{li2018harmonious} adds a scaling layer for learning a fusion scale to combine the channel attention and spatial attention outputs. In this paper, we adopt a feature extractor with a normalizer for all multi-level activations to deal with the bias.

%\paragraph{Dynamic filter networks}
Our proposed method is also related with the dynamic filter networks~\cite{jia2016dynamic} in which sample-specific filter parameters are generated based on the input. The filters with a specific kernel size are dynamically produced to enable local spatial transformations on the input feature map, which serve as dynamic convolution kernels. Comparatively, in our method, the adaptive fusion weights are applied on the multi-level response maps to obtain a desired fused output and a feature extractor with a normalizer is proposed to alleviate the bias during  training.

\section{Feature Fusion with Fixed Weights}
Before expounding the proposed model, we first introduce the notations used in our model and revisit the fixed weight fusion in CASENet~\cite{yu2017casenet} as preliminaries.

%\subsection{Fixed Weight Fusion in CASENet}

CASENet adopts ResNet-101 with some minor modifications to extract multi-level features. Based on the backbone, a $1 \times 1$ convolution layer and a following upsampling layer are connected to the output of the first three and the top stacks of residual blocks, producing three single channel feature maps $\{A_{side1}, A_{side2}, A_{side3}\}$ and a $K$-channel class activation map $A_{side5}=\{A_{side5}^1,...,A_{side5}^K\}$ respectively. Here $K$ is the number of categories. The shared concatenation replicates the bottom features $\{A_{side1}, A_{side2}, A_{side3}\}$ for $K$ times to separately concatenate with each of the $K$ top activations in $A_{side5}$:
\begin{align}
A_{cat} &=\{A_{cat}^1,...,A_{cat}^K\},\\
A_{cat}^i &=\{A_{side5}^i, A_{side1}, A_{side2}, A_{side3}\},i\in[1,K].
\end{align}
The resulting concatenated activation map $A_{cat}$ is then fed into the $K$-grounped 1 $\times$ 1 conv layer to produce the fused activation map with $K$ channels:
\begin{align}
A_{fuse}^i &\!=\!w_1^i A_{side5}^i\!+\! w_2^i A_{side1} \!+\! w_3^i A_{side2} \!+\! w_4^i A_{side3} \label{eq.3} \\
A_{fuse} &=\{A_{fuse}^i\},i\in[1,K]
\end{align}
where $(w_1^i,w_2^i,w_3^i,w_4^i)$ are the parameters of the $K$-grounped 1 $\times$ 1 conv layer, serving as fusion weights for the $i$th category. We omit the bias term in Eqn. (\ref{eq.3}) for simplicity. One can refer to ~\cite{yu2017casenet} for more details.

%\subsection{Discussion}
CASENet imposes the same fixed set of fusion weights $(w_1^i,w_2^i,w_3^i,w_4^i)$ of the $i$th category upon all the images across all the locations of the feature maps. By reproducing CASENet, we make following observations. i) The weight $w_1^i$ always surpasses  the other  weights $(w_2^i,w_3^i,w_4^i)$ for each class significantly. ii) By evaluating the performance of $A_{side5}$ and the fused output $A_{fuse}$, we find they give almost the same performance. This implies  that the low-level feature responses $\{A_{side1}, A_{side2}, A_{side3}\}$ contribute very little to the final fused output although they contain fine-grained edge location and structure information. The final decision is mainly determined by high-level features. As such, the CASENet cannot exploit low-level information sufficiently and produce only coarse edges.

\section{Proposed Model}
To address the above limitations of existing \textit{Semantic Edge Detection} (SED) models in fusing features with fixed weights, we develop a new SED model with dynamic feature fusion. The proposed model fuses multi-level features through two modules: 1) the feature extractor with a normalizer is to normalize the  magnitude scales of multi-level features and 2) the adaptive weight fusion module is to learn adaptive fusion weights for different locations of multi-level feature maps (see Figure~\ref{fig2} (a)).
\subsection{Dynamic Feature Fusion}
The first module is inspired by  following observations.
In CASENet~\cite{yu2017casenet},
%with the supervision imposed on Side5 and the final fused activation,
the edge responses  from the top layers are much stronger than the ones from other three bottom outputs. Such variation in scales of activations  biases the multi-level feature fusion to the responses from the top layers. Moreover,  the top-layer output is much similar to the ground-truth for the training examples,  due to the direct  supervision.  Applying the same weights to all the locations forces the network to learn much higher fusion weights for the top-level features, undesirably ignoring contributions from low-level features.
This further inhibits the low-level features from providing fine edge information for detecting object boundaries in the cases of heavily biased distributions of edge and non-edge pixels.

Therefore, before feature fusion, we first deal with the scale variation of multi-level responses by normalizing their magnitudes. In this way, the following  adaptive weight learner can get rid of distractions from scale variation and learn effective fusion weights more easily.

%\textcolor{red}{[need to explain how to do the normalization.]}
A feature extractor with a normalizer (see Figure ~\ref{fig2} (a)) is employed to normalize multi-level responses to the similar magnitude. Concretely, side feature normalization blocks in the module are responsible for normalizing feature maps of the corresponding level.

To achieve our proposed dynamic feature fusion, we devise two different schemes for predicting the adaptive fusion weights, \emph{i.e.}, location-invariant and location-adaptive fusion weights (see Figure ~\ref{fig2} (b)). The former treats all locations in the feature maps equally and universal fusion weights are learned according to specific input adaptively. The later one adjusts the fusion weights conditioned on location features of the image and   lifts  the contributions of low-level features to locating fine edges along  object boundaries.

Concretely, given the activation of the multi-level side outputs $A_{side}$ of size $H \times W$, we aim to obtain a fused output $A_{fuse}=f(A_{side})$ by aggregating multi-level responses. The fusion operation in CASENet in Eqn.~\eqref{eq.3} can be written as the following generalized form:
\begin{equation}
A_{fuse}=f(A_{side};\textbf{W})
\end{equation}
where $f$ encapsulates operations in Eqn.~\eqref{eq.3} and $\textbf{W}=(w_1^i,w_2^i,w_3^i,w_4^i)$ denotes the fusion weights.

Differently, we propose an adaptive weight learner to actively learn the fusion weights conditioned on the feature map itself, which is formulated as
\begin{equation}
A_{fuse}=f(A_{side};\Psi(x)),
\end{equation}
where $x$ denotes the feature map. The above formulations characterize the essential difference between our proposed adaptive weight fusion method and the fixed weight fusion method. We enforce the fusion weights $\textbf{W}$ to explicitly depend on the feature map $x$, \emph{i.e.} $\textbf{W}=\Psi(x)$. Different input feature maps $x$ would induce different parameters $\Psi(x)$ and thus lead to modifications to the adaptive weight learner $f(.;.)$ dynamically. In this way, the semantic edge detection model can fast adapt to the input image and favourably learn proper multi-level response fusion weights in an end-to-end manner.

Regarding $\Psi(x)$, corresponding to the two fusion weight schemes,  there are two   adaptive weight learners, namely location-invariant weight learner and location-adaptive weight learner. The location-invariant weight learner  learns $4K$ fusion weights in total as shown in the following equation, which are shared by all locations of feature maps to be fused:
\begin{equation}
\Psi(x)=(w_1^i, w_2^i,w_3^i,w_4^i),i\in[1,K].
\end{equation}
However, the location-adaptive weight learner generates $4K$ fusion weights for each spatial location, which results in $4KHW$ weighting parameters in total.
\begin{align}
\Psi(x) &=(\mathbf{w_{s,t}}),s\in[1,H],t\in[1,W] \\
\mathbf{w_{s,t}} &=((w_1^i)_{s,t}, (w_2^i)_{s,t},(w_3^i)_{s,t},(w_4^i)_{s,t}),i\in[1,K]
\end{align}
The location-invariant weight learner generates universal fusion weights for all locations, while the location-adaptive weight learner tailors  fusion weights for each location considering the spatial variety.
%QH: Actually I do not quite understand your explanations in subsection 4.1. Lack motivations.

\subsection{Network Architecture}

Our proposed network architecture is  based on ResNet~\cite{he2016deep} and adopts the same modifications as CASENet~\cite{yu2017casenet}  to preserve low-level edge information, as shown in Figure 2. Side feature normalization blocks are connected to the first three and the fifth stack of residual blocks. This block consists of a $1 \times 1$ convolution layer, a Batch Normalization (BN)~\cite{ioffe2015batch} layer and a deconvolution layer. The $1 \times 1$ convolution layer produces single and $K$ channel response maps for Side1-3 and Side5 respectively. The BN layer is applied on the output of the $1 \times 1$ convolution layer to normalize the multi-level responses to the same magnitude. Deconvolution layers are then used  to upsample the response maps to the original image size.

Another side feature normalization block is connected to the fifth stack of the residual block, in which a $4K$-channel feature map is produced. The adaptive weight learner then takes in the output of the Side5-w feature normalization block to predict the dynamic fusion weights $w(x)$. We design two instantiations for location-invariant weight learner and location-adaptive weight learner respectively, as detailed below.

\paragraph{Location-invariant weight learner} Location-invariant weight learner (see Figure 2 (b)) is a na\"ive version of adaptive weight learner. The output of the Side5-w feature normalization block $x$ with size $H \times W \times 4K$ is taken as the input of a global average pooling layer to produce a $4K$-channel vector. After that, {a block of alternatively connected FC, BN and ReLU layers} is repeated twice and followed by {FC and BN layers} to generate the $4K$ location-invariant fusion weights for the fused response map $A_{fuse}$ with size $H \times W \times 4K$. These fusion weights are conditional on the input content but all locations share the same fusion parameters.

\paragraph{Location-adaptive weight learner} We further propose the location-adaptive weight learner (see Figure 2 (b)) to solve the shortcomings of fixed weight fusion. The feature map $x$ is forwarded to {a block of alternatively connected $1 \times 1$ conv, BN and ReLU layers},  repeated twice and followed by a $1 \times 1$ {conv and BN layer}, to generate location-aware fusion weights $w(x)$ with size $H \times W \times 4K$. Element-wise multiplication and category-wise summation are then applied to the location-aware fusion weights and the fused response map $A_{fuse}$ to generate the final fused output. The location-adaptive weight learner predicts $4K$ fusion weights for each spatial location on Side1-3 and Side5 {to allow low-level responses to provide fine  edge locations for object boundaries.} One may consider applying softmax activation function to the output of the adaptive weight leaner to learn fusion weights mutually-exclusive for the side feature maps to be fused respectively. The ablation experiments in Section 5.2 demonstrate that {the activation function hampers the performance in the case of generating adaptive fusion weights}.

\iffalse
\begin{table*}[t]
\centering
\resizebox{\textwidth}{!}{
\begin{tabular}{c|c|c|c|c|c|c|c|c|c|c|c|c|c|c|c|c|c|c|c|c|c||c}
\toprule
Method & Activation & BN & road & side. &buil. & wall & fence & pole & light & sign & vege. & terrain & sky & person & rider & car & truck & bus & train & mot. & bike & mean \\
\midrule \midrule
DFF-sigmoid & sigmoid & \checkmark & 93.4 & 86.2 & \textbf{89.6} & 57.6 & 63.4 & 84.8 & 85.0 & \textbf{84.3} & \textbf{92.2} & 71.8 & 92.4 & 91.0 & 82.2 & 94.8 & 55.2 & 74.7 & 57.6 & 74.3 & 86.5 & 79.8 \\
DFF-softmax & softmax & \checkmark & \textbf{93.6} & \textbf{86.5} & \textbf{89.6} & \textbf{58.9} & 63.7 & 84.9 & 84.9 & 84.1 & 92.1 & \textbf{72.0} & \textbf{92.6} & 90.8 & 82.2 & 94.5 & 58.0 & 76.0 & \textbf{58.8} & 70.9 & \textbf{87.0} & 80.1 \\
DFF-w/o-BN & - & - & 93.1 & 86.3 & 89.4 & 57.0 & 63.5 & 84.6 & 85.3 & 84.2 & 92.0 & 71.1 & 92.3 & 90.6 & 81.3 & 94.5 & 56.9 & 72.2 & 48.4 & 70.1 & 86.5 & 78.9 \\
DFF & - & \checkmark & 93.3 & \textbf{86.5} & 89.5 & 58.2 & \textbf{64.3} & \textbf{85.2} & \textbf{85.5} & \textbf{84.3} & \textbf{92.2} & \textbf{72.0} & \textbf{92.6} & \textbf{91.2} & \textbf{82.6} & \textbf{94.9} & \textbf{60.0} & \textbf{77.2} & 56.9 & \textbf{74.5} & 86.3 & \textbf{80.4} \\
\bottomrule
\end{tabular}}
\caption{The effectiveness of each component in our proposed model on Cityscapes dataset. All MF scores are measured by \%}
\label{tab:table2}
\end{table*}
\fi

\begin{table}[h]
\centering
\resizebox{0.48\textwidth}{!}{
\begin{tabular}{c|c|cc|c|cc}
\toprule
\multirow{2}*{Method} & \multirow{2}*{Normalizer} & \multicolumn{3}{c|}{Learner} & \multirow{2}*{mean} & \multirow{2}*{$\Delta$ mean} \\
\cline{3-5}
&  & invariant & adaptive & softmax \\
\midrule
Baseline            &            &  &  &  & 78.4 & \\
\midrule
\multirow{5}*{Ours}
& \checkmark &              &               &               & 78.7          & $+$ 0.3 \\ % \cline{2-7}
& \checkmark & \checkmark   &               &               & 79.1          & $+$ 0.7 \\ % \cline{2-7}
& \checkmark &              & \checkmark    & \checkmark    & 80.1          & $+$ 1.7 \\
& \checkmark &              & \checkmark    &               & \textbf{80.4} & $+$ \textbf{2.0} \\ % \cline{2-7}
&            &              & \checkmark    &               & 78.9          & $+$ 0.5 \\
\bottomrule
\end{tabular}}
\vspace{-2mm}
\caption{The effectiveness of each component in our proposed model on Cityscapes dataset. The mean value of MF scores over all categories is measured by \%.}
\vspace{-4mm}
\label{tab:table1}
\end{table}

\begin{table*}[t]
\centering
\resizebox{\textwidth}{!}{
\begin{tabular}{c|c|c|c|c|c|c|c|c|c|c|c|c|c|c|c|c|c|c|c|c|c||c}
\toprule
Method & backbone & crop size & road & side. &buil. & wall & fence & pole & light & sign & vege. & terrain & sky & person & rider & car & truck & bus & train & mot. & bike & mean \\
\midrule \midrule
\multirow{3}*{DFF}
%& ResNet50 & 472 & 200 & 93.0 & 86.2 & 88.7 & 59.1 & 63.0 & 85.1 & 83.7 & 83.1 & 91.6 & 71.5 & 92.0 & 90.5 & 81.2 & 94.5 & 48.9 & 65.2 & 38.1 & 69.2 & 85.9 & 77.4 \\
& ResNet50 & 512 & 93.1 & \textbf{86.5} & 89.0 & 58.8 & 63.5 & 84.8 & 83.7 & 83 & 91.8 & 71.9 & 92.2 & 90.7 & 81.3 & 94.7 & 58.5 & 72.7 & 50.8 & 71.7 & 86.6 & 79.2 \\
& ResNet50 & 640 & \textbf{93.3} & \textbf{86.5} & 89.5 & 58.2 & \textbf{64.3} & 85.2 & \textbf{85.5} & \textbf{84.3} & \textbf{92.2} & \textbf{72.0} & \textbf{92.6} & \textbf{91.2} & 82.6 & \textbf{94.9} & 60.0 & 77.2 & 56.9 & \textbf{74.5} & 86.3 & 80.4 \\
& ResNet101 & 512 & 93.2 & 86.3 & \textbf{89.7} & \textbf{60.0} & \textbf{64.3} & \textbf{86.0} & 85.2 & 83.7 & 92.0 & 71.3 & 92.3 & 91.1 & \textbf{83.4} & 94.8 & \textbf{64.5} & \textbf{77.4} & \textbf{59.9} & 72.1 & \textbf{86.9} & \textbf{80.7} \\
\bottomrule
\end{tabular}
}
\vspace{-2mm}
\caption{Results of the proposed DFF using different backbone networks and crop sizes. All MF scores are measured by \%}
\vspace{-3mm}
\label{tab:table2}
\end{table*}

\begin{table*}[t]
\centering
\resizebox{\textwidth}{!}{
\begin{tabular}{c|c|c|c|c|c|c|c|c|c|c|c|c|c|c|c|c|c|c|c||c}
\toprule
Method & road & side. &buil. & wall & fence & pole & light & sign & vege. & terrain & sky & person & rider & car & truck & bus & train & mot. & bike & mean \\
\midrule \midrule
CASENet & 86.6 & 78.8 & 85.1 & 51.5 & 58.9 & 70.1 & 70.8 & 74.6 & 83.5 & 62.9 & 79.4 & 81.5 & 71.3 & 86.9 & 50.4 & 69.5 & 52.0 & 61.3 & 80.2 & 71.3 \\
DDS-R & 90.5 & 84.2 & 86.2 & 57.7 & 61.4 & 85.1 & 83.8 & 80.4 & 88.5 & 67.6 & 88.2 & 89.9 & 80.1 & 91.8 & 58.6 & 76.3 & 56.2 & 68.8 & \textbf{87.3} & 78.0 \\
%DFF & 93.3 & \textbf{86.5} & 89.5 & \textbf{58.2} & \textbf{64.3} & \textbf{85.2} & \textbf{85.5} & \textbf{84.3} & \textbf{92.2} & \textbf{72.0} & \textbf{92.6} & \textbf{91.2} & \textbf{82.6} & \textbf{94.9} & \textbf{60.0} & \textbf{77.2} & \textbf{56.9} & \textbf{74.5} & \textbf{86.3} & \textbf{80.4} \\
DFF & \textbf{93.2} & \textbf{86.3} & \textbf{89.7} & \textbf{60.0} & \textbf{64.3} & \textbf{86.0} & \textbf{85.2} & \textbf{83.7} & \textbf{92.0} & \textbf{71.3} & \textbf{92.3} & \textbf{91.1} & \textbf{83.4} & \textbf{94.8} & \textbf{64.5} & \textbf{77.4} & \textbf{59.9} & \textbf{72.1} & 86.9 & \textbf{80.7} \\
\midrule \midrule
SEAL (0.0035) & 87.6 & 77.5 & 75.9 & 47.6 & 46.3 & 75.5 & 71.2 & 75.4 & 80.9 & 60.1 & 87.4 & 81.5 & 68.9 & 88.9 & 50.2 & 67.8 & 44.1 & 52.7 & 73.0 & 69.1 \\
DFF (0.0035) & \textbf{89.4} & \textbf{80.1} & \textbf{79.6} & \textbf{51.3} & \textbf{54.5} & \textbf{81.3} & \textbf{81.3} & \textbf{81.2} & \textbf{83.6} & \textbf{62.9} & \textbf{89.0} & \textbf{85.4} & \textbf{75.8} & \textbf{91.6} & \textbf{54.9} & \textbf{73.9} & \textbf{51.9} & \textbf{64.3} & \textbf{76.4} & \textbf{74.1} \\
\bottomrule
\end{tabular}
}
\vspace{-2mm}
\caption{Comparison with state-of-the-arts on Cityscapes dataset. All MF scores are measured by \%}
\vspace{-3mm}
\label{tab:table3}
\end{table*}

\begin{table*}[t]
\centering
\resizebox{\textwidth}{!}{
\begin{tabular}{c|c|c|c|c|c|c|c|c|c|c|c|c|c|c|c|c|c|c|c|c||c}
\toprule
Method & aer. & bike & bird & boat & bottle & bus & car & cat & chair & cow & table & dog & horse & mot. & per. & pot. & sheep & sofa & train & tv & mean \\
\midrule \midrule
CASENet & 83.6 & 75.3 & 82.3 & 63.1 & 70.5 & 83.5 & 76.5 & 82.6 & 56.8 & 76.3 & 47.5 & 80.8 & 80.9 & 75.6 & 80.7 & 54.1 & 77.7 & 52.3 & 77.9 & 68.0 & 72.3 \\
DDS-R & 85.4 & 78.3 & 83.3 & 65.6 & 71.4 & 83.0 & 75.5 & 81.3 & \textbf{59.1} & 75.7 & 50.7 & 80.2 & 82.7 & 77.0 & 81.6 & \textbf{58.2} & 79.5 & 50.2 & 76.5 & 71.2 & 73.3 \\
SEAL & 84.5 & 76.5 & 83.7 & 64.9 & 71.7 & 83.8 & 78.1 & 85.0 & 58.8 & 76.6 & \textbf{50.9} & 82.4 & 82.2 & 77.1 & 83.0 & 55.1 & 78.4 & \textbf{54.4} & 79.3 & 69.6 & 73.8 \\
\midrule
DFF & \textbf{86.5} & \textbf{79.5} & \textbf{85.5} & \textbf{69.0} & 73.9 & \textbf{86.1} & \textbf{80.3} & \textbf{85.3} & 58.5 & \textbf{80.1} & 47.3 & \textbf{82.5} & \textbf{85.7} & \textbf{78.5} & \textbf{83.4} & 57.9 & \textbf{81.2} & 53.0 & \textbf{81.4} & \textbf{71.6} & \textbf{75.4} \\
\bottomrule
\end{tabular}
}
\vspace{-2mm}
\caption{Comparison with state-of-the-arts on SBD dataset. All MF scores are measured by \%}
\vspace{-3mm}
\label{tab:table4}
\end{table*}

\begin{figure*}[t]
\centering
\resizebox{\textwidth}{!}{
\includegraphics[]{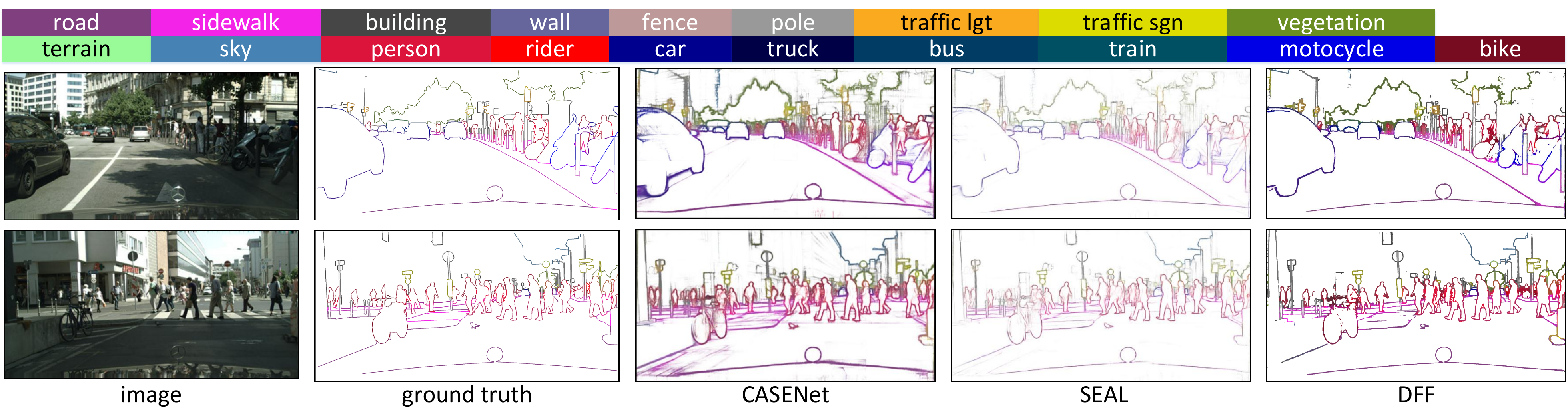}
}
%\vspace{-3mm}
\caption{Qualitative comparison on Cityscapes among ground truth, CASENet, SEAL and DFF (ordering from left to right in the figure). Best viewed in color.}
%\vspace{-3mm}
\label{fig3}
\end{figure*}

\section{Experiments}

\subsection{Experimental Setup}

\paragraph{Datasets and augmentations} We evaluate our proposed method on two popular benchmarks for semantic edge detection: Cityscapes~\cite{cordts2016cityscapes} and SBD~\cite{hariharan2011semantic}.  During training, we use random mirroring and random cropping on both Cityscapes and SBD. For Cityscapes, we also augment each training sample with random scaling in $[0.75, 2]$, and the base size for random scaling is set as 640. For SBD, we augment the training data by resizing each image with scaling factors $\{0.5, 0.75, 1, 1.25, 1.5\}$ following~\cite{yu2017casenet}. During testing, the images with original image size are used for Cityscapes and the images are padded to $512 \times 512$ for SBD.

\paragraph{Training strategies} The proposed network is built on ResNet~\cite{he2016deep} pretrained on ImageNet~\cite{imagenet_cvpr09}.
The network is trained with standard reweighted cross-entropy loss~\cite{yu2017casenet} and optimized by SGD using PyTorch~\cite{paszke2017automatic}\footnote{Codes will be released soon.}. We set the base learning rate to 0.08 and 0.05 for Cityscapes and SBD respectively; the ``poly" policy is used for learning rate decay. The crop size, batch size, training epoch, momentum, weight decay are set to $640 \times 640$ / $352 \times 352$, 8 / 16, 200 / 10, 0.9, $1e-4$ respectively for Cityscapes and SBD. All experiments are performed using 4 NVIDIA TITAN Xp(12GB) GPUs with synchronized batch normalization~\cite{Zhang_2018_CVPR}. We use the same data augmentation and hyper-parameters for reproducing CASENet and training the proposed model for fair comparison.

\paragraph{Evaluation protocol} We follow~\cite{yu2018simultaneous} to evaluate the performance with stricter rules than the benchmark used in~\cite{yu2017casenet,liu2018semantic}. The ground truth maps are downsampled into half size of original dimensions for Cityscapes, and are generated with instance-sensitive edges for both datasets. Maximum F-measure (MF) at optimal dataset scale (ODS) with matching distance tolerance set as 0.02 is used for Cityscapes and SBD evaluation following our baseline methods. For  fair comparison with~\cite{yu2018simultaneous}, we also set the distance tolerance to 0.0035 for Cityscapes dataset.

\subsection{Ablation Study}

We first conduct ablation experiments on Cityscapes to investigate the effectiveness of each newly proposed module. We use CASENet as the baseline model and add our proposed normalizer and learner on it. All the ablation experiments use the same settings with 640 $\times$ 640 crop size and ResNet50 backbone. We report the reproduced baseline CASENet trained with the same training settings for fair comparison, which actually has higher accuracy than the original paper. Results are summarized in Table~\ref{tab:table1}.

\paragraph{Normalizer} We first evaluate the effect of the proposed normalizer in the feature extractor. As shown in Table~\ref{tab:table1} (row 2), the normalizer provides 0.3\% performance gain compared with the baseline which does not normalize the multi-level features to the same magnitude before fusion. It shows that normalized features are more suitable for multi-level information fusion than the unnormalized ones. This is probably because the unnormalized features have dramatically different magnitudes, making the learning process biased to the features with higher magnitude.

\paragraph{Location-invariant learner}
Row 3 in Table~\ref{tab:table1} shows the results of further adding the proposed location-invariant learner which adaptively fuses multi-level features conditioned on the input image. As shown in Table~\ref{tab:table1}, the location-invariant learner provides another 0.4\% performance gain compared with \textit{baseline+normalizer} that assigns fixed fusion weights without considering the uniqueness of each input image. The performance boost successfully shows that different input images with different illumination, semantics, \emph{etc.}, may require different fusion weights and applying universal weights to all training images will lead to less satisfactory results.

\paragraph{Location-adaptive learner}
The location-adaptive learner  provides location-aware fusion weights that are conditioned on not only the input image but also the local context, which further improves the location-invariant learner. As shown in Table~\ref{tab:table1} row 4 to row 5, location-adaptive learner improves the performance significantly from 79.1\% to 80.4\%, indicating a fusion module should consider both the uniqueness of each input image and the local context at each location. The proposed fusion module in DFF successfully captures both of them and thus delivers remarkable performance gain. Besides, it can be seen that the location-adaptive learner heavily relies on pre-normalized multi-level features, since the performance degrades by 1.5\% when the proposed normalizer is removed from the framework, as shown in row 5 and row 6. This further justifies the effectiveness of the proposed normalizer for rescaling the activation values to the same magnitude before conducting fusion.

\paragraph{Activation function} Row 4 and row 5 in Table~\ref{tab:table1} show the cases of further constraining the generated fusion weights for better performance. In particular, we append a softmax layer to the end of location-adaptive weight learner (Figure~\ref{fig2} (b)) to force the learned fusion weights for each category to add up to 1, and compare its results with the unconstrained version. However, we observe performance drop of 0.3\% with softmax activation function adopted. Our intuitive explanation is that applying softmax activation function to the fusion weights normalizes their values to the range $[0,1]$, which may actually be negative for some response maps or some locations. Furthermore, adopting softmax activation function enforces the fusion weights of  multi-level response maps for each category to be mutually exclusive, which makes the response maps compete with each other to be more important. However, it is desirable that each response map be given enough freedom to be active and be useful to the final fused output. Therefore, we do not use any activation functions to constrain the fusion weights in the proposed DFF model.

\paragraph{Deeper backbone network} We also perform experiments on deeper backbones to verify our model's generalization. As shown in Table~\ref{tab:table2}, we evaluate our model performance on ResNet101 with $512\times512$ crop size. Note that we do not use $640\times640$, simply because of the GPU memory limitation. In order to make a fair comparison, we also train a DFF on ResNet50 with $512\times512$ crop size. The result shows DFF also generalizes well on deep ResNet101 and outperforms its counterpart (ResNet50, $512 \times 512$) by 1.5\%. Higher performance might be achieved if we can train the ResNet101 with $640 \times 640$ crop size using GPUs with larger memory, \textit{e.g.} Nvidia V100 (32GB).

\subsection{Comparison with State-of-the-arts}
We compare the performance of the proposed DFF model with state-of-the-art semantic edge detection methods~\cite{yu2017casenet,yu2018simultaneous,liu2018semantic} on Cityscapes dataset and SBD dataset.

\subsubsection{Results on Cityscapes}

We compare the proposed DFF model with CASENet and DDS, and evaluate MF (ODS) with matching distance tolerance set as 0.02. In addition, the matching distance tolerance is decreased to 0.0035 to make fair comparison with SEAL. As can be seen in Table~\ref{tab:table3}, our DFF model outperforms all these well established baselines and achieves new state-of-the-art 80.7\% MF (ODS). The DFF model is superior to other models for most classes. Specifically, the MF (ODS) of the proposed DFF model is 9.4\% higher than CASENet and 2.7\% higher than DDS on average. It is also worth noting that DFF achieves 5\% higher than SEAL under stricter matching distance tolerance (0.0035), which reflects the  DFF model can locate more accurate edges by taking full advantage of multi-level response maps, especially the low-level ones, with dynamic feature fusion. We also visualize some prediction results for qualitative comparison shown in Figure~\ref{fig3}. Through comparing with CASENet and SEAL, we can observe DFF can predict more accurate and clearer edges on object boundaries while suppressing the fragmentary and trivial edge responses inside the objects.

\subsubsection{Results on SBD}

Compared with Cityscapes, SBD dataset has fewer objects and especially fewer overlapping ones in each image. In addition, it contains many misaligned annotations. We compare the proposed DFF model with state-of-the-arts on SBD validation set in Table~\ref{tab:table4}. The proposed DFF model outperforms all the baselines and achieves new state-of-the-art 75.4\% MF (ODS), which well confirms its superiority.

\section{Conclusion}

In this paper, we propose a novel dynamic feature fusion (DFF) model to exploit multi-level responses for producing a desired fused output. The newly introduced normalizer scales multi-level side activations to the same magnitude preparing for the down-streaming fusion operation. Location-adaptive weight learner is proposed to actively learn customized fusion weights conditioned on the individual sample content for each spatial location. Comprehensive experiments on Cityscapes and SBD demonstrate the proposed DFF model can improve the performance by locating more accurate edges on object boundaries as well as suppressing trivial edge responses inside the objects. In the future, we will continue to improve the location-adaptive weight learner by considering both the high-level and low-level feature maps from the backbone.

\bibliographystyle{named}
\bibliography{ijcai19}

\end{document}